\documentclass{article}

\usepackage[final, nonatbib]{neurips_2022}

\usepackage[utf8]{inputenc} 
\usepackage[T1]{fontenc}    
\usepackage{hyperref}       
\usepackage{url}            
\usepackage{booktabs}       
\usepackage{amsfonts}       
\usepackage{nicefrac}       
\usepackage{microtype}      
\usepackage{xcolor}         

\usepackage{cite}
\usepackage{graphicx}
\usepackage{pifont}
\usepackage{xspace}

\makeatletter
\DeclareRobustCommand\onedot{\futurelet\@let@token\@onedot}
\def\@onedot{\ifx\@let@token.\else.\null\fi\xspace}

\def\eg{\emph{e.g}\onedot} 
\def\ie{\emph{i.e}\onedot}

\def\wrt{w.r.t\onedot} 

\makeatother

\newcommand{\cmark}{\ding{51}}
\newcommand{\xmark}{\ding{55}}

\title{FQDet: Fast-converging Query-based Detector}

\author{%
  Cédric Picron \\
  ESAT-PSI, KU Leuven \\
  \texttt{cedric.picron@esat.kuleuven.be} \\
  \And
  Punarjay Chakravarty \\
  Ford Greenfield Labs, Palo Alto \\
  \texttt{pchakra5@ford.com} \\
  \And
  Tinne Tuytelaars \\
  ESAT-PSI, KU Leuven \\
  \texttt{tinne.tuytelaars@esat.kuleuven.be} \\
}

\begin{document}

\maketitle

\begin{abstract}
    Recently, two-stage Deformable DETR introduced the query-based two-stage head, a new type of two-stage head different from the region-based two-stage heads of classical detectors as Faster R-CNN. In query-based two-stage heads, the second stage selects one feature per detection processed by a transformer, called the query, as opposed to pooling a rectangular grid of features processed by CNNs as in region-based detectors. In this work, we improve the query-based head by improving the prior of the cross-attention operation with anchors, significantly speeding up the convergence while increasing its performance. Additionally, we empirically show that by improving the cross-attention prior, auxiliary losses and iterative bounding box mechanisms typically used by DETR-based detectors are no longer needed. By combining the best of both the classical and the DETR-based detectors, our FQDet head peaks at $45.4$ AP on the 2017 COCO validation set when using a ResNet-50+TPN backbone, only after training for 12 epochs using the 1x schedule. We outperform other high-performing two-stage heads such as \eg Cascade R-CNN, while using the same backbone and while being computationally cheaper. Additionally, when using the large ResNeXt-101-DCN+TPN backbone and multi-scale testing, our FQDet head achieves $52.9$ AP on the 2017 COCO test-dev set after only 12 epochs of training. Code is released at \url{https://github.com/CedricPicron/FQDet}.
\end{abstract}

\section{Introduction}
Deep neural networks designed for solving the object detection task, are commonly subdivided in a \textit{backbone} and a \textit{head}. The backbone is defined as taking in an image and outputting a set of feature maps. These feature maps are typically of different resolutions, with a factor two separating consecutive maps in width and height, hence forming what is known as a \textit{feature pyramid} \cite{lin2017feature}. A common choice for the backbone is using a ResNet-50 \cite{he2016deep} network in combination with a pyramid network (PN) such as FPN \cite{lin2017feature}, PANet \cite{liu2018path}, BiFPN \cite{tan2020efficientdet} or TPN \cite{picron2021trident}.

The object detection head is defined as taking in a set of feature maps from the backbone, and outputting object detection predictions, with each prediction consisting of an axis-aligned box together with a corresponding class label and confidence score. Object detection heads are commonly subdivided in \textit{one-stage} and \textit{two-stage} heads. One-stage heads make an object detection prediction for \textit{every feature} from its input set of feature maps. Two-stage heads first evaluate for every feature whether it is related to an object, and then use these results to \textit{focus the processing} on object-related parts of the feature maps.

In what follows, we exclusively work and compare with two-stage heads. Two-stage heads can further be subdivided in \textit{region-based} and \textit{query-based} two-stage heads, based on how the second stage of the two-stage head is implemented. 

Region-based two-stage heads pool a rectangular grid of features from the backbone feature maps and further process these using convolutional neural networks (CNNs) in order to obtain their final object predictions. Region-based heads are found in many well-established two-stage object detectors such as Faster R-CNN \cite{ren2015faster} and Cascade R-CNN \cite{cai2018cascade}.

Query-based two-stage heads were recently introduced in the two-stage variant of Deformable DETR \cite{zhu2020deformable}. While region-based heads extract a \textit{grid of features} from the backbone feature maps per detection, only a \textit{single feature} called the query is selected in query-based heads. These queries are then further processed using operations commonly found in a transformer decoder \cite{vaswani2017attention}, namely cross-attention, self-attention, and feedforward operations. Here the cross-attention operation is defined between the backbone feature maps and each query, the self-attention operation is applied between the queries from different predictions, and the feedforward operations are performed on each query individually.

In this paper, we propose a new query-based two-stage head, called the FQDet (Fast-converging Query-based Detector) head. FQDet was obtained by combining the query-based paradigm from two-stage Deformable DETR \cite{zhu2020deformable} with classical object detection techniques such as anchor generation and static (\ie non-Hungarian) matching.

When evaluated on the 2017 COCO object detection benchmark \cite{lin2014microsoft} after training for $12$ epochs, our FQDet head with ResNet-50+TPN backbone achieves $45.4$ AP, outperforming other high-performing two-stage heads such as Faster R-CNN \cite{ren2015faster}, Cascade R-CNN \cite{cai2018cascade}, two-stage Deformable DETR \cite{zhu2020deformable} and Sparse R-CNN \cite{sun2021sparse} when using the same backbone.

\section{Related work}

\paragraph{DETR and its variants.} With its unique way of approaching the object detection task, DETR \cite{carion2020end} quickly gained a lot of popularity. Since, many variants such as SMCA \cite{gao2021fast}, Conditional DETR \cite{meng2021conditional}, Deformable DETR \cite{zhu2020deformable}, Anchor DETR \cite{wang2022anchor}, DAB-DETR \cite{liu2022dab} and DN-DETR \cite{li2022dn} have been proposed to improve the main two shortcomings of DETR, namely its slow convergence speed and its poor performance on small objects. In doing so, Deformable DETR \cite{zhu2020deformable} introduces the query-based two-stage head, a new type of two-stage head now also used in other works such as DINO \cite{zhang2022dino}.

\paragraph{Anchors.} Our FQDet head differs from other query-based two-stage heads by introducing anchors within the query-based two-stage paradigm. Faster R-CNN \cite{ren2015faster} was one of the first works to use anchors. Anchors are axis-aligned boxes of different sizes and aspect ratios that are attached to backbone feature locations and are refined to yield the final bounding box predictions. Anchors are found in both one-stage detectors (SSD \cite{liu2016ssd}, YOLO \cite{redmon2016you}, RetinaNet \cite{lin2017focal}) and two-stage detectors (Faster R-CNN \cite{ren2015faster}, Cascade R-CNN \cite{cai2018cascade}). Over the years, many anchor-free detectors have also been proposed such as CornerNet \cite{law2018cornernet}, FCOS \cite{tian2019fcos}, CenterNet \cite{zhou2019objects}, FoveaBox \cite{kong2020foveabox}, DETR \cite{carion2020end} and Deformable DETR \cite{zhu2020deformable}.

\paragraph{Matching.} Our FQDet head additionally differs from other query-based two-stage heads by using a static top-k matching scheme as opposed to the dynamic Hungarian matching scheme. Matching is the process responsible of assigning ground-truth labels to the different model outputs during training. Most detectors have a \textit{static} matching scheme, meaning that the matching scheme stays the same throughout the whole training process. Static matching schemes are common when using anchors, where label assignment is done based on the IoU overlap between anchors and ground-truth boxes (\eg Faster R-CNN \cite{ren2015faster}, RetinaNet \cite{lin2017focal} and Cascade R-CNN \cite{cai2018cascade}). Some works make use of a \textit{dynamic} matching scheme, meaning that the matching scheme changes as training progresses. In Dynamic R-CNN \cite{zhang2020dynamic}, the minimum IoU thresholds are increased throughout the training process. In DETR \cite{carion2020end}, a dynamic Hungarian matching scheme is used, matching every ground-truth to a single prediction (one-to-one matching). In OTA \cite{ge2021ota}, the one-to-one Hungarian matching scheme from DETR is extended to one-to-many Hungarian matching, where a single ground-truth could be assigned to multiple predictions.

\section{Method}

\subsection{Overview and motivation} \label{sec:overview_motiv}
\paragraph{Overview.} Below, we give an overview of the main improvements of our FQDet head over two-stage Deformable DETR (see Appendix~\ref{sec:revisit} for more information about DETR and Deformable DETR):
\begin{enumerate}
    \item We use multiple anchors of different sizes and aspect ratios \cite{lin2017focal}, improving the cross-attention prior.
    \item We encode the bounding boxes relative to these anchors, similar to Faster R-CNN \cite{ren2015faster}.
    \item We only use an L1 loss for bounding box regression \cite{ren2015faster} without GIoU loss \cite{rezatofighi2019generalized}.
    \item We remove the auxiliary decoder losses and predictions as used in DETR \cite{carion2020end}.
    \item We do not perform iterative bounding box regression as in Deformable DETR \cite{zhu2020deformable}.
    \item We replace Hungarian matching \cite{carion2020end, zhu2020deformable} with static top-k matching (\textit{ours}).
\end{enumerate}

\paragraph{Motivation.} In what follows, we motivate each of the above FQDet design choices. The impact of these design choices on the performance will be analyzed in the ablation studies of Subsection~\ref{sec:base_abl}.

\textit{(Item 1)} We first motivate the use of anchors in our FQDet head. To understand this, we must take a look at the cross-attention operation from Deformable DETR. This operation updates a query based on features sampled from the backbone feature maps, where the sample coordinates are regressed from the query itself. These sample coordinates are defined \wrt to a reference frame based on the query box prior, where the origin of the reference frame is placed at the center of the box prior and where the unit lengths correspond to half the width and height of the box prior. By using anchors of various sizes and aspect ratios in our FQDet head, we significantly improve these cross-attention box priors resulting in more accurate and robust sampling, and eventually in better performance.

Items 2-5 from above list now in fact all follow from the use of anchors. \textit{(Item 2)} Given that we use anchors, it is logical to also encode our bounding box predictions relative to these anchors as \eg done in Faster R-CNN \cite{ren2015faster}. \textit{(Item 3)} Since we encode our boxes relative to the anchors, we only use an L1 loss (without GIoU loss) for the bounding box regression, as commonly used for this box encoding \cite{ren2015faster}. \textit{(Item 4)} Since we improved our box priors with the introduction of anchors, there is no need to further improve these by introducing intermediate box predictions supervised by auxiliary losses as Deformable DETR \cite{zhu2020deformable}. In our setting, updating the box priors with these intermediate box predictions even hurts the performance, as a fixed sample reference frame for the different decoder layers is to be preferred over a changing one. \textit{(Item 5)} Given that we do not make intermediate box predictions, there is also no need for iterative bounding box regression as used in Deformable DETR \cite{zhu2020deformable}.

\textit{(Item 6)} Our final design choice involves replacing the Hungarian matching with our static top-k matching. At the beginning of training, Hungarian matching might assign a ground-truth detection to a query which sampled from a very different region compared to the location of the ground-truth detection. Hungarian matching hence produces many low-quality matches at the beginning of training, significantly slowing down convergence. Instead, we propose our static (\ie non-Hungarian) top-k matching scheme consisting of matching each ground-truth detection with its top-k anchors. Queries corresponding to these anchors are then also automatically matched. This matching scheme guarantees that each matched query indeed did processing in the neighborhood of the matched ground-truth detection.

\subsection{FQDet in detail}
\begin{figure}
    \centering
    \includegraphics[scale=0.95]{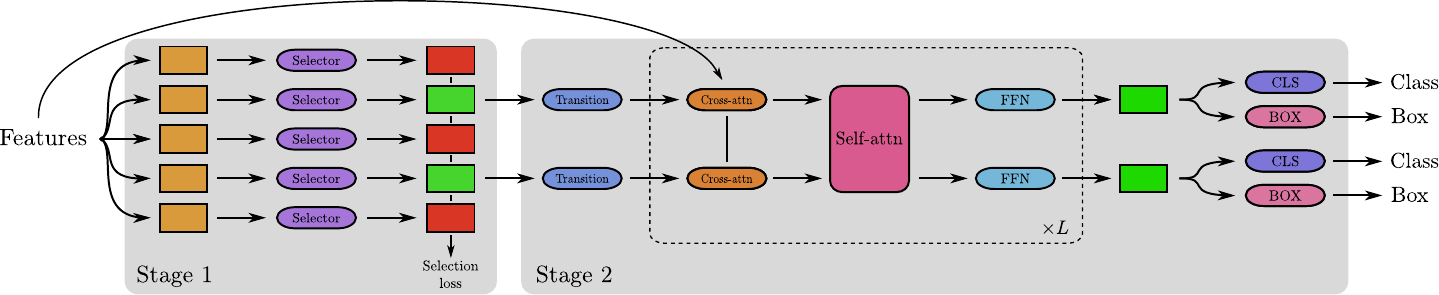}
    \caption{Architectural overview of our two-stage FQDet head. At its input, the head receives a set of features from the backbone, for which it is currently unknown whether they belong to an object or not (orange boxes). In Stage~1, a lightweight selector network is used to determine which features belong to an object (green boxes) and which do not (red boxes). Features predicted to belong to an object are selected for Stage~2 and are now called queries. In Stage~2, queries first pass through a lightweight transition network, before being processed by $L$ consecutive transformer decoder layers. The resulting queries (flashy green boxes) are then processed by the classification and bounding box regression networks to yield the object detection predictions.}
    \label{fig:FQDet}
    \vspace{-0.4cm}
\end{figure}

An architectural overview of our FQDet head is displayed in Figure~\ref{fig:FQDet}. The head consists of two stages: a first stage applied to all input features and a second stage applied only to those features (\ie queries) selected from Stage~1. We now discuss both stages in more depth.

\paragraph{Stage 1.} In Stage~1, the goal of our FQDet head is to determine which input features belong to an object, such that these can be selected for Stage~2. At its input, the head receives a set of backbone feature maps in the form of a feature pyramid. To each of those backbone feature locations we attach anchors of various sizes and aspect ratios. A lightweight selector network is then applied to each backbone feature outputting a score for each feature-anchor combination, with top-scoring feature-anchor combinations selected for Stage~2. In Stage~2, the selected feature becomes the query and the corresponding anchor becomes the query box prior for the cross-attention operation.

A selection loss is used to learn which features-anchor combinations should be selected. The selection targets are determined by our static top-k matching scheme, where anchors with a top-k highest IoU with one of the ground-truth boxes receive a positive label and anchors with no top-k highest IoU with any ground-truth box receive a negative label.

\paragraph{Stage 2.} In Stage~2, the goal of our FQDet head is to make an object detection prediction for each of the queries selected from Stage~1. First, the queries are passed through a lightweight transition network. This transition network is meant to change the queries from the backbone feature space to the head feature space, as
both feature spaces might have a different size.

Once processed by the transition network, the queries are passed through $L$ consecutive transformer decoder layers. Each layer consists of a sequence of cross-attention, self-attention and a feedforward network (FFN). The \textit{cross-attention} operation updates each query by sampling features from the backbone feature maps using the multi-scale deformable attention (MSDA) operation from Deformable DETR \cite{zhu2020deformable}. The \textit{self-attention} operation updates each query through interaction with other queries using a standard multi-head attention (MHA) operation \cite{vaswani2017attention}. Finally, each query is also processed individually by a \textit{feedforward network} (FFN) which is a specific type of MLP \cite{vaswani2017attention}.

After the decoder layers, the queries are processed by the classification and bounding box regression networks yielding the object detection predictions. The classification network outputs a score for each of the classes within the dataset, along with a score for the non-object class. The bounding box regression network outputs the bounding box deltas \wrt the anchor, using the same box encoding scheme as in Faster R-CNN \cite{ren2015faster}. Both classification and bounding box regression networks are implemented using lightweight MLPs. 

We use the same static top-k matching scheme in Stage~2 as in Stage~1, except that we allow for a different $k$ in both stages.

\section{Experiments} \label{sec:experiments}
\subsection{Setup} \label{sec:setup}

\paragraph{Dataset.} We perform our experiments on the COCO detection dataset \cite{lin2014microsoft}, where we train on the 2017 COCO training set and evaluate on the 2017 COCO validation or test-dev set.

\paragraph{Implementation details.} Throughout our experiments, we use a ResNet-50+TPN backbone. For the ResNet-50 network \cite{he2016deep}, we use ImageNet \cite{deng2009imagenet} pretrained weights provided by TorchVision and freeze the stem, stage~1 and BatchNorm layers (see \cite{radosavovic2020designing} for the used terminology). For the TPN network \cite{picron2021trident}, we use the TPN configuration with $3$ TPN layers and $2$ bottleneck layers per self-processing operation. The TPN network outputs a feature pyramid containing five feature maps, ranging from $P_3$ to $P_7$, each having a feature size of $256$. Here, feature map $P_i$ is downscaled $i$ times with a factor $2$ in both width and height.

For the selector network of our FQDet head, we use one hidden layer and one output layer. The hidden layer consists of a bottleneck layer \cite{he2016deep} with the same settings as used in TPN \cite{picron2021trident}. The output layer is a linear projection layer projecting each feature to a selection score for each of the anchor types. We use $9$ anchor types of different size and aspect ratio as defined in RetinaNet \cite{lin2017focal}. For the selection loss, we use a sigmoid focal loss \cite{lin2017focal} with $\alpha=0.25$ and $\gamma=2.0$ as hyperparameters. The selection targets are determined using our static top-k matching scheme with $k = 5$.

The $300$ feature-anchor combinations (\ie query and query prior boxes) with the highest score are selected for the second stage. The queries pass first through one of the $9$ transition networks, depending on their corresponding anchor type. Each of these transition networks consist of a linear projection layer with output feature size $256$, preceded by a LayerNorm and a ReLU activation layer.

The transformer decoder of the FQDet head consists of $6$ consecutive decoder layers, with each layer a sequence of cross-attention, self-attention and feedforward operations. Each of the operations updates the queries by using skip connections, with each residual branch starting with a LayerNorm layer.

The cross-attention operation is implemented by the multi-scale deformable attention (MSDA) operation from Deformable DETR \cite{zhu2020deformable}. We use $8$ MSDA attention heads and $4$ MSDA sampling points per head and feature pyramid level. For the self-attention operation, we use the multi-head attention operation (MHA) from \cite{vaswani2017attention} with $8$ attention heads. The feedforward network (FFN) is implemented as in \cite{vaswani2017attention} with a hidden feature size of $1024$, except that we remove the dropout layers.

Additionally, anchor encodings are added to the queries when passed to the residual branches of the cross-attention and self-attention operations. These anchor encodings could be interpreted as enhanced positional encodings containing an additional notion of width and height. The anchor encodings are learned from the anchor boxes written in (center location, width-height) format, using an MLP with one hidden layer.

For the FQDet classification and bounding box regression networks, we use in both cases an MLP with one hidden layer. The classification and bounding box ground-truth labels are obtained using our static top-k matching scheme with $k=15$. Unmatched queries receive the \textit{non-object} label. For the classification loss, we use a sigmoid focal loss \cite{lin2017focal} with the same hyperparameters as for the selection loss. We use an L1 loss for the bounding box regression in the delta box encoding space as done in Faster R-CNN \cite{ren2015faster}. We remove duplicate detections during inference by using non-maximum suppression (NMS) with an IoU threshold of $0.5$.

We train our models with the AdamW optimizer \cite{loshchilov2017decoupled} with weight decay $10^{-4}$. We use an initial learning rate of $10^{-5}$ for the backbone parameters and for the linear projection modules computing the MSDA sampling offsets. For the remaining model parameters, we use an initial learning rate of $10^{-4}$. Our results are obtained by using the 1x training schedule, consisting of $12$ training epochs with learning rate drops after the $9$th and $11$th epoch with a factor $0.1$. Our models are trained using our internal cluster on $2$ GPUs with batch size $2$ each, while using the same data augmentation scheme as in DETR \cite{carion2020end}.

\paragraph{Evaluation.} When evaluating a model (\ie backbone with head), we will report both its object detection performance as well as its computational characteristics. For the object detection performance, we report the Average Precision (AP) metrics as provided by the COCO benchmark \cite{lin2014microsoft}. For the computational characteristics, we use following five metrics:

\begin{enumerate}
    \item The number of model parameters (Params).
    \item The number of training FPS (tFPS).
    \item The peak training memory utilization (tMem).
    \item The number of inference FPS (iFPS).
    \item The peak inference memory utilization (iMem).
\end{enumerate}

The FPS and memory metrics are obtained on a GeForce GTX 1660 Ti GPU by supplying the model with a batch of two $800 \times 800$ images, each containing $10$ ground-truth objects during training. The FPS metric should be interpreted as the number of times the GPU can process above input per second. Note that both forward and backward passes (with parameter update from the optimizer) are used to obtain the training characteristics.

\subsection{Base FQDet results and ablation studies} \label{sec:base_abl}
In Table~\ref{tab:FQDet_base}, we report the results of our base FQDet head (\ie our FQDet head with the settings explained above). Our base FQDet head obtains $44.3$ AP after only $12$ epochs of training, outperforming the two-stage Deformable DETR head by $5.4$ AP (see Subsection~\ref{sec:comp}). 

We now perform a series of ablation studies investigating to what extent each of the design choices as listed in Subsection~\ref{sec:overview_motiv}, contribute to the increased performance of FQDet.

\begin{table*}[t!]
    \centering
    \caption{Base FQDet results on the 2017 COCO validation set.}
    \resizebox{\columnwidth}{!}{
        \begin{tabular}{c c|c c c c c c|c c c c c}
            \toprule
            Head & Epochs & AP & AP$_{50}$ & AP$_{75}$ & AP$_S$ & AP$_M$ & AP$_L$ & Params & tFPS & tMem & iFPS & iMem \\ \midrule
            FQDet & $12$ & $44.3$ & $62.5$ & $47.8$ & $25.3$ & $48.3$ & $59.0$ & $42.6$ M & $1.6$ & $2.86$ GB & $4.8$ & $0.52$ GB \\
            \bottomrule
        \end{tabular}}
    \label{tab:FQDet_base}
\end{table*}

\begin{table*}[t!]
    \centering
    \caption{FQDet anchors ablation study on the 2017 COCO validation set.}
    \resizebox{\columnwidth}{!}{
        \begin{tabular}{c c c|c c c c c c|c c c c c}
            \toprule
            Head & Sizes & Ratios & AP & AP$_{50}$ & AP$_{75}$ & AP$_S$ & AP$_M$ & AP$_L$ & Params & tFPS & tMem & iFPS & iMem \\ \midrule
            FQDet & $1$ & $1$ & $40.6$ & $60.8$ & $42.4$ & $24.1$ & $44.6$ & $53.4$ & $42.0$ M & $1.7$ & $2.84$ GB & $5.0$ & $0.51$ GB \\
            FQDet & $3$ & $1$ & $42.9$ & $61.4$ & $46.2$ & $24.4$ & $47.3$ & $57.5$ & $42.2$ M & $1.7$ & $2.84$ GB & $5.0$ & $0.51$ GB \\
            FQDet & $1$ & $3$ & $43.1$ & $62.6$ & $46.1$ & $25.8$ & $46.8$ & $57.0$ & $42.2$ M & $1.7$ & $2.84$ GB & $5.0$ & $0.51$ GB \\
            FQDet & $3$ & $3$ & $44.3$ & $62.5$ & $47.8$ & $25.3$ & $48.3$ & $59.0$ & $42.6$ M & $1.6$ & $2.86$ GB & $4.8$ & $0.52$ GB \\
            \bottomrule
        \end{tabular}}
    \label{tab:anchors}
\end{table*}

\paragraph{Anchors ablation study.} In the anchors ablation study, we vary the number of anchor types used in the FQDet selector. In our base FQDet head, we use $9$ anchor types composed of $3$ anchor sizes and $3$ anchor aspect ratios. In Table~\ref{tab:anchors}, we show the FQDet results when the number of anchor sizes and/or the number of anchor aspect ratios is reduced to $1$. We can see that the number of anchor types play an important role in the resulting performance, with a performance drop up to $3.7$ AP when only one anchor type is used. Moreover, we observe that the additional computational cost of using multiple anchor types is very limited. These results support our analysis from Subsection~\ref{sec:overview_motiv} where we highlight the importance of having good query box priors (\ie anchors) for the cross-attention operation.

\begin{table*}[t!]
    \centering
    \caption{FQDet box loss ablation study on the 2017 COCO validation set.}
    \resizebox{\columnwidth}{!}{
        \begin{tabular}{c c c|c c c c c c|c c c c c}
            \toprule
            Head & L1 & GIoU & AP & AP$_{50}$ & AP$_{75}$ & AP$_S$ & AP$_M$ & AP$_L$ & Params & tFPS & tMem & iFPS & iMem \\ \midrule
            FQDet & \cmark & \cmark & $43.0$ & $60.3$ & $46.2$ & $23.5$ & $46.9$ & $58.1$ & $42.6$ M & $1.6$ & $2.86$ GB & $4.8$ & $0.52$ GB \\
            FQDet & \cmark & \xmark & $44.3$ & $62.5$ & $47.8$ & $25.3$ & $48.3$ & $59.0$ & $42.6$ M & $1.6$ & $2.86$ GB & $4.8$ & $0.52$ GB \\
            \bottomrule
        \end{tabular}}
    \label{tab:box_loss}
\end{table*}

\paragraph{Box loss ablation study.} In the box loss ablation study, we compare the L1 bounding box loss used by our base FQDet head, with a weighted combination of L1 and GIoU losses as used in DETR \cite{carion2020end} and Deformable DETR \cite{zhu2020deformable}. Note that for the weighted combination of L1 and GIoU losses, we also change the classification weight from $1.0$ to $2.0$, in order to be consistent with Deformable DETR. The results of the box loss ablation study are found in Table~\ref{tab:box_loss}. We can see that using a simple L1 box loss outperforms the weighted combination of L1 and GIoU losses by $1.3$ AP. As argued in Subsection~\ref{sec:overview_motiv}, we believe an L1 loss suffices when encoding the bounding boxes relative to anchors as done in FQDet.

\begin{table*}[t!]
    \centering
    \caption{FQDet auxiliary losses ablation study on the 2017 COCO validation set.}
    \resizebox{\columnwidth}{!}{
        \begin{tabular}{c c c|c c c c c c|c c c c c}
            \toprule
            Head & Aux. & Shared & AP & AP$_{50}$ & AP$_{75}$ & AP$_S$ & AP$_M$ & AP$_L$ & Params & tFPS & tMem & iFPS & iMem \\ \midrule
            FQDet & \cmark & \cmark & $44.1$ & $62.4$ & $47.5$ & $25.2$ & $48.0$ & $59.3$ & $42.6$ M & $1.5$ & $2.88$ GB & $4.5$ & $0.52$ GB \\
            FQDet & \cmark & \xmark & $43.7$ & $62.1$ & $46.7$ & $25.3$ & $47.6$ & $58.4$ & $43.5$ M & $1.5$ & $2.89$ GB & $4.5$ & $0.52$ GB \\
            FQDet & \xmark & $-$ & $44.3$ & $62.5$ & $47.8$ & $25.3$ & $48.3$ & $59.0$ & $42.6$ M & $1.6$ & $2.86$ GB & $4.8$ & $0.52$ GB \\
            \bottomrule
        \end{tabular}}
    \label{tab:aux}
\end{table*}

\paragraph{Auxiliary losses ablation study.} In the auxiliary losses ablation study, we investigate whether making intermediate predictions after each decoder layer supervised by auxiliary losses as in Deformable DETR \cite{zhu2020deformable}, improves the performance of our base FQDet head. In Table~\ref{tab:aux}, we show the performance of our FQDet head where auxiliary losses are added, both with and without shared classification and bounding box regression networks. We observe that the performance decreases between $0.2$ and $0.6$ AP, depending on whether the regression networks are shared or not. We argue that as our FQDet head already provides good box priors through its Stage~1 anchors for the Stage~2 decoder, there is no need to further improve these through intermediate box predictions supervised by auxiliary losses.

\begin{table*}[t!]
    \centering
    \caption{FQDet IBBR ablation study on the 2017 COCO validation set.}
    \resizebox{\columnwidth}{!}{
        \begin{tabular}{c c c c|c c c c c c|c c c c c}
            \toprule
            Head & Aux. & IBBR & Shared & AP & AP$_{50}$ & AP$_{75}$ & AP$_S$ & AP$_M$ & AP$_L$ & Params & tFPS & tMem & iFPS & iMem \\ \midrule
            FQDet & \cmark & \cmark & \cmark & $43.5$ & $62.0$ & $46.5$ & $24.7$ & $47.6$ & $58.1$ & $42.6$ M & $1.5$ & $2.89$ GB & $4.4$ & $0.52$ GB \\
            FQDet & \cmark & \cmark & \xmark & $43.8$ & $62.2$ & $46.7$ & $24.7$ & $47.7$ & $58.8$ & $43.5$ M & $1.5$ & $2.90$ GB & $4.4$ & $0.52$ GB \\
            FQDet & \xmark & \xmark & $-$ & $44.3$ & $62.5$ & $47.8$ & $25.3$ & $48.3$ & $59.0$ & $42.6$ M & $1.6$ & $2.86$ GB & $4.8$ & $0.52$ GB \\
            \bottomrule
        \end{tabular}}
    \label{tab:ibbr}
\end{table*}

\paragraph{IBBR ablation study.} In the iterative bounding box regression (IBBR) ablation study, we again investigate the setting with intermediate predictions supervised by auxiliary losses as in previous ablation study, except that we also do iterative bounding box regression (IBBR) as introduced in Deformable DETR \cite{zhu2020deformable}. IBBR consists in iteratively updating the bounding box predictions relative to the previous bounding box predictions (or anchors). The results of our IBBR ablation study are found in Table~\ref{tab:ibbr}, where we again consider the cases with and without shared classification and bounding box regression networks. We can see that the performance again decreases compared to our base FQDet head, this time between $0.5$ and $0.8$ AP depending on whether the regression networks were shared or not. Hence even when adding IBBR to the setting of intermediate predictions supervised by auxiliary losses, does not yield improvements for our FQDet head.

\begin{table*}[t!]
    \centering
    \caption{FQDet matching ablation study on the 2017 COCO validation set.}
    \resizebox{\columnwidth}{!}{
        \begin{tabular}{c c|c c c c c c|c c c c c}
            \toprule
            Head & Matching & AP & AP$_{50}$ & AP$_{75}$ & AP$_S$ & AP$_M$ & AP$_L$ & Params & tFPS & tMem & iFPS & iMem \\ \midrule
            FQDet & Hungarian & $35.1$ & $54.2$ & $37.5$ & $20.3$ & $38.5$ & $43.8$ & $42.6$ M & $1.5$ & $2.86$ GB & $4.8$ & $0.52$ GB \\
            FQDet & Absolute & $43.3$ & $61.0$ & $46.8$ & $25.5$ & $47.4$ & $57.6$ & $42.6$ M & $1.7$ & $2.86$ GB & $4.8$ & $0.52$ GB \\
            FQDet & Top-k (\textit{ours}) & $44.3$ & $62.5$ & $47.8$ & $25.3$ & $48.3$ & $59.0$ & $42.6$ M & $1.6$ & $2.86$ GB & $4.8$ & $0.52$ GB \\
            \bottomrule
        \end{tabular}}
    \label{tab:matching}
\end{table*}

\paragraph{Matching ablation study.} In the matching ablation study, we compare our static top-k matching scheme, with the static absolute matching scheme from Faster R-CNN \cite{ren2015faster} and the Hungarian matching scheme from Deformable DETR \cite{zhu2020deformable} (Stage~2 only). In Table~\ref{tab:matching}, we show the results when using our FQDet head with each of these three matching schemes. We can see that our top-k matching scheme works best, outperforming the absolute matching scheme with $1.0$ AP and the Hungarian scheme with $9.2$ AP. These 12-epoch results nicely support the analysis from Subsection~\ref{sec:overview_motiv}, where we argue that Hungarian matching leads to slow convergence due to the low-quality matches at the beginning of training.

\begin{table*}[t!]
    \centering
    \caption{Improving the base FQDet head by changing the number of MSDA sampling points, altering the inference strategy and training longer (evaluated on 2017 COCO validation set).}
    \vspace{0.1cm}
    \resizebox{\columnwidth}{!}{
        \begin{tabular}{c c c c|c c c c c c|c c c c c}
            \toprule
            Head & Points & Inference & Epochs & AP & AP$_{50}$ & AP$_{75}$ & AP$_S$ & AP$_M$ & AP$_L$ & Params & tFPS & tMem & iFPS & iMem \\ \midrule
            FQDet & $4$ & Old & $12$ & $44.3$ & $62.5$ & $47.8$ & $25.3$ & $48.3$ & $59.0$ & $42.6$ M & $1.6$ & $2.86$ GB & $4.8$ & $0.52$ GB \\
            FQDet & $1$ & Old & $12$ & $44.5$ & $62.9$ & $47.7$ & $26.1$ & $48.7$ & $59.3$ & $42.0$ M & $1.7$ & $2.84$ GB & $5.0$ & $0.51$ GB \\
            FQDet & $1$ & New & $12$ & $45.4$ & $64.3$ & $48.9$ & $27.4$ & $49.6$ & $60.3$ & $42.0$ M & $1.7$ & $2.84$ GB & $5.0$ & $0.51$ GB \\
            FQDet & $1$ & New & $24$ & $46.2$ & $65.3$ & $49.2$ & $29.1$ & $50.3$ & $61.0$ & $42.0$ M & $1.7$ & $2.84$ GB & $5.0$ & $0.51$ GB \\
            \bottomrule
        \end{tabular}}
    \label{tab:improvements}
\end{table*}

\subsection{Improving the base FQDet head} \label{sec:improving}
In what follows, we make following three improvements to the base FQDet head. The results after applying each of those improvements sequentially, are shown in Table~\ref{tab:improvements}.

\paragraph{MSDA sampling points improvement.} The first improvement consists in changing the number of MSDA sampling points per head and per feature pyramid level from $4$ to $1$, increasing the performance by $0.2$ AP while being computationally cheaper.

\paragraph{Inference strategy improvement.} The second improvement consists in altering the inference strategy. In the current inference strategy, we only assign a single classification label to each predicted bounding box, \ie the label with the highest score. This is sub-optimal as the head might be hesitant about the class, with multiple classes containing similar scores. In the new inference strategy, we therefore consider all classification labels with their scores for each predicted bounding box, and feed these to the NMS algorithm to only keep the top $100$ predictions. Replacing the old inference strategy with the new one improves the performance by $0.9$ AP, while leaving the computational cost virtually unchanged.

\paragraph{Training longer.} The performance of our FQDet head can further be improved by training for $24$ epochs using the 2x schedule, increasing the performance by $0.8$ AP.

\begin{table*}[t!]
    \centering
    \caption{Comparison of our FQDet head with other prominent two-stage heads on the 2017 COCO validation set.}
    \vspace{0.1cm}
    \resizebox{\columnwidth}{!}{
        \begin{tabular}{c c|c c c c c c|c c c c c}
            \toprule
            Head & Epochs & AP & AP$_{50}$ & AP$_{75}$ & AP$_S$ & AP$_M$ & AP$_L$ & Params & tFPS & tMem & iFPS & iMem \\ \midrule
            Faster R-CNN & $12$ & $40.7$ & $60.9$ & $44.4$ & $23.9$ & $46.0$ & $54.1$ & $50.3$ M & $1.8$ & $2.62$ GB & $4.0$ & $0.55$ GB \\
            Cascade R-CNN & $12$ & $43.9$ & $61.9$ & $47.6$ & $25.8$ & $49.3$ & $58.5$ & $77.9$ M & $1.4$ & $3.15$ GB & $2.5$ & $0.65$ GB \\
            Deformable DETR & $12$ & $38.9$ & $58.1$ & $41.6$ & $23.6$ & $41.7$ & $52.3$ & $43.0$ M & $1.6$ & $3.02$ GB & $4.9$ & $0.52$ GB \\
            Sparse R-CNN & $12$ & $39.8$ & $57.4$ & $43.1$ & $23.8$ & $42.5$ & $53.4$ & $114.8$ M & $1.3$ & $4.67$ GB & $4.0$ & $0.79$ GB \\
            \midrule
            FQDet (ours) & $12$ & $\mathbf{45.4}$ & $\mathbf{64.3}$ & $\mathbf{48.9}$ & $\mathbf{27.4}$ & $\mathbf{49.6}$ & $\mathbf{60.3}$ & $42.0$ M & $1.7$ & $2.84$ GB & $5.0$ & $0.51$ GB \\
            \bottomrule
        \end{tabular}}
    \label{tab:comparison}
\end{table*}

\subsection{Comparison with other two-stage heads} \label{sec:comp}
In Table~\ref{tab:comparison}, we compare our improved FQDet head from Subsection~\ref{sec:improving} with other prominent two-stage heads from the literature. We compare with the two-stage heads from Faster R-CNN \cite{ren2015faster}, Cascade R-CNN \cite{cai2018cascade}, Deformable DETR \cite{zhu2020deformable} and Sparse R-CNN \cite{sun2021sparse}, while using the same ResNet-50+TPN backbone and training settings. We used their MMDetection implementations \cite{mmdetection} where we adapted the config files to be compatible with the $P_3$ to $P_7$ feature pyramid outputted by the backbone.

From Table~\ref{tab:comparison}, we see the clear superiority of our FQDet head. Our FQDet head outperforms all other two-stage heads with $1.5$ up to $6.5$ AP, not only outperforming the query-based two-stage Deformable DETR head, but also well-established region-based heads such as Cascade R-CNN. Additionally, our FQDet head is considerably cheaper compared to most of the heads, only being slightly more expensive during training than the Faster R-CNN head.

\begin{figure}
    \centering
    \begin{minipage}{0.49\textwidth}
        \centering
        \includegraphics[scale=0.42]{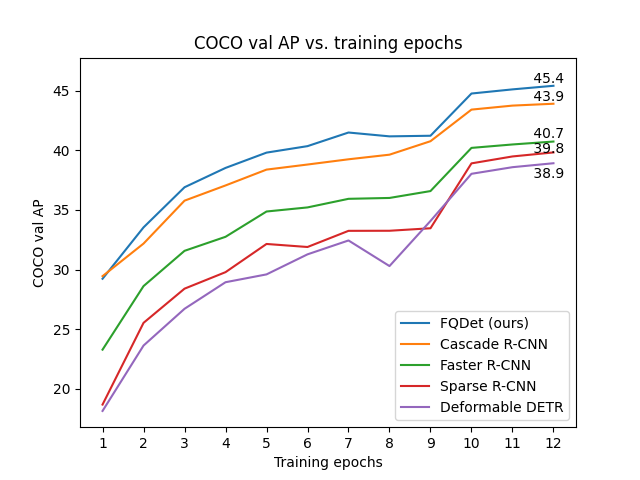}
    \end{minipage} \hfill
    \begin{minipage}{0.49\textwidth}
        \centering
        \includegraphics[scale=0.42]{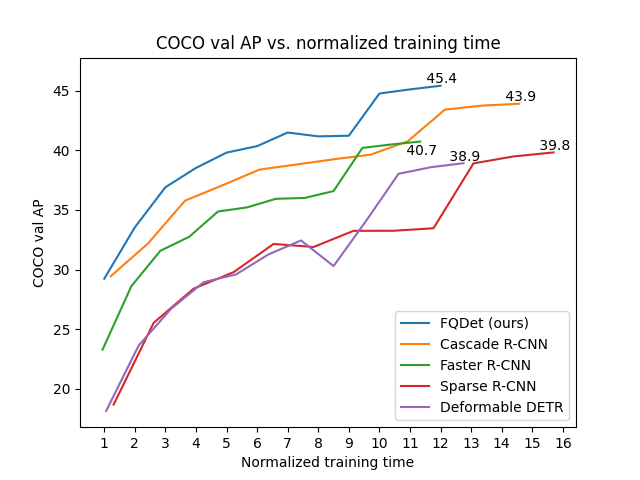}
    \end{minipage}
    \caption{(\textit{Left}) Convergence graph displaying the COCO validation AP after every training epoch for different two-stage heads. (\textit{Right}) Convergence graph displaying the COCO validation AP as a function of normalized training time. Here one unit of normalized training time is defined as the time it takes to train our FQDet head for one epoch.}
    \label{fig:two-stage-comp}
    \vspace{-0.4cm}
\end{figure}

In Figure~\ref{fig:two-stage-comp}, we show two convergence graphs containing our FQDet two-stage head and other prominent two-stage heads using their MMDetection \cite{mmdetection} implementation. Here the left convergence graph displays the COCO validation AP after every training epoch, whereas the right convergence graph shows the COCO validation AP after each unit of normalized time, with this unit of time defined as the time it takes to train our FQDet head for one epoch.

When looking at the left convergence graph, we can see that our FQDet head performs better throughout the whole training process, highlighting its fast convergence. Only the Cascade R-CNN head trails by a small amount. It takes however substantially more time for the Cascade R-CNN head to complete one epoch compared to our FQDet head. When taking the actual training time into account in the right convergence graph, we can now see that our FQDet performs better compared to the Cascade R-CNN at any point in time during the training process. Our FQDet hence demonstrates excellent performance achieved in few training epochs, while remaining relatively cheap. 

\subsection{Comparison with state-of-the-art methods}
In this subsection, we compare our FQDet detector with other state-of-the-art object detection methods, when using the ResNeXt-101-64$\times$4d-DCN backbone \cite{xie2017aggregated, dai2017deformable, zhu2019deformable}. To further boost performance, we make following modifications: 1) we consider a $P_2$ to $P_7$ feature pyramid, 2) we replace convolutions with modulated deformable convolutions \cite{zhu2019deformable} in TPN, 3) we use group convolutions \cite{xie2017aggregated} with $32$ groups in TPN, and 4) we consider the $(L, B) = (3, 4)$ TPN configuration \cite{picron2021trident}. Our test-time augmentation (TTA) scheme consists of horizontal flipping and multi-scale testing with the shorter edges of images rescaled to $[500, 600, 700, 800, 900, 1000, 1100, 1200, 1300]$.

\begin{table*}[t!]
    \centering
    \caption{Comparison of FQDet with state-of-the-art object detection methods on the 2017 COCO test-dev set. All methods use the ResNeXt-101-64$\times$4d-DCN backbone. The `TTA' column indicates whether test-time augmentations such as horizontal flipping and multi-scale testing are used.}
    \vspace{0.1cm}
    \resizebox{0.85\columnwidth}{!}{
        \setlength{\tabcolsep}{7pt}
        \begin{tabular}{c c c|c c c c c c}
            \toprule
            Method & Epochs & TTA & AP & AP$_{50}$ & AP$_{75}$ & AP$_S$ & AP$_M$ & AP$_L$ \\ \midrule
            ATSS \cite{zhang2020bridging} & $24$ & \xmark & $47.7$ & $66.5$ & $51.9$ & $29.7$ & $50.8$ & $59.4$ \\
            Sparse R-CNN \cite{sun2021sparse} & $36$ & \xmark & $48.9$ & $68.3$ & $53.4$ & $29.9$ & $50.9$ & $62.4$ \\
            Deformable DETR \cite{zhu2020deformable} & $50$ & \xmark & $50.1$ & $69.7$ & $54.6$ & $30.6$ & $52.8$ & $64.7$ \\
            TOOD \cite{feng2021tood} & $24$ & \xmark & $51.1$ & $69.4$ & $55.5$ & $31.9$ & $54.1$ & $63.7$ \\
            RelationNet++ \cite{chi2020relationnet++} & $24$ & \xmark & $50.3$ & $69.0$ & $55.0$ & $32.8$ & $55.0$ & $65.8$ \\
            DyHead \cite{dai2021dynamic} & $24$ & \xmark & $52.3$ & $70.7$ & $57.2$ & $35.1$ & $56.2$ & $63.4$ \\
            \textbf{FQDet} (ours) & $12$ & \xmark & $51.5$ & $70.3$ & $55.7$ & $31.9$ & $54.5$ & $65.5$ \\
            \midrule
            ATSS \cite{zhang2020bridging} & $24$ & \cmark & $50.7$ & $68.9$ & $56.3$ & $33.2$ & $52.9$ & $62.4$ \\
            Sparse R-CNN \cite{sun2021sparse} & $36$ & \cmark & $51.5$ & $71.1$ & $57.1$ & $34.2$ & $53.4$ & $64.1$ \\
            Deformable DETR \cite{zhu2020deformable} & $50$ & \cmark & $52.3$ & $71.9$ & $58.1$ & $34.4$ & $54.4$ & $65.6$ \\
            RelationNet++ \cite{chi2020relationnet++} & $24$ & \cmark & $52.7$ & $70.4$ & $58.3$ & $35.8$ & $55.3$ & $64.7$ \\
            DyHead \cite{dai2021dynamic} & $24$ & \cmark & $\mathbf{54.0}$ & $72.1$ & $\mathbf{59.3}$ & $\mathbf{37.1}$ & $\mathbf{57.2}$ & $\mathbf{66.3}$ \\
            \textbf{FQDet} (ours) & $12$ & \cmark & $52.9$ & $\mathbf{72.3}$ & $57.2$ & $34.6$ & $55.8$ & $\mathbf{66.3}$ \\
            \bottomrule
        \end{tabular}}
    \label{tab:sota}
\end{table*}

In Table~\ref{tab:sota}, we compare our FQDet object detector with other state-of-the-art detectors, all using the ResNeXt-101-64$\times$4d-DCN backbone. We can see that our FQDet detector outperforms all other detectors, both with and without test-time augmentations, except for the DyHead detector \cite{dai2021dynamic}.

\subsection{Limitations} \label{sec:limitations}
Only a single dataset (\ie COCO \cite{lin2014microsoft}) was used to evaluate our method. Experiments were only run a single time as running experiments multiple times would be too computationally expensive.

\subsection{Potential negative societal impacts} \label{sec:impact}
Object detection models could be used in military applications to \eg identify human targets and could be used for mass surveillance.

\section{Conclusion}
We propose the FQDet head, a new query-based two-stage object detection head combining the best of both classical and DETR-based detectors. Our FQDet head improves the cross-attention prior with anchors and introduces the effective top-k matching scheme. We perform various ablation studies validating the FQDet design choices and show improvements \wrt existing two-stage heads when compared on the COCO object detection benchmark after training for $12$ epochs.

\section*{Acknowledgements}
This work was supported by the Ford-KU Leuven alliance program.

\bibliography{egbib}

\begin{thebibliography}{10}

\bibitem{cai2018cascade}
Zhaowei Cai and Nuno Vasconcelos.
\newblock Cascade r-cnn: Delving into high quality object detection.
\newblock In {\em Proceedings of the IEEE conference on computer vision and
  pattern recognition}, pages 6154--6162, 2018.

\bibitem{carion2020end}
Nicolas Carion, Francisco Massa, Gabriel Synnaeve, Nicolas Usunier, Alexander
  Kirillov, and Sergey Zagoruyko.
\newblock End-to-end object detection with transformers.
\newblock In {\em European Conference on Computer Vision}, pages 213--229.
  Springer, 2020.

\bibitem{mmdetection}
Kai Chen, Jiaqi Wang, Jiangmiao Pang, Yuhang Cao, Yu~Xiong, Xiaoxiao Li,
  Shuyang Sun, Wansen Feng, Ziwei Liu, Jiarui Xu, Zheng Zhang, Dazhi Cheng,
  Chenchen Zhu, Tianheng Cheng, Qijie Zhao, Buyu Li, Xin Lu, Rui Zhu, Yue Wu,
  Jifeng Dai, Jingdong Wang, Jianping Shi, Wanli Ouyang, Chen~Change Loy, and
  Dahua Lin.
\newblock {MMDetection}: Open mmlab detection toolbox and benchmark, 2019.
\newblock URL: \url{https://github.com/open-mmlab/mmdetection}.

\bibitem{chi2020relationnet++}
Cheng Chi, Fangyun Wei, and Han Hu.
\newblock Relationnet++: Bridging visual representations for object detection
  via transformer decoder.
\newblock {\em Advances in Neural Information Processing Systems},
  33:13564--13574, 2020.

\bibitem{dai2017deformable}
Jifeng Dai, Haozhi Qi, Yuwen Xiong, Yi~Li, Guodong Zhang, Han Hu, and Yichen
  Wei.
\newblock Deformable convolutional networks.
\newblock In {\em Proceedings of the IEEE international conference on computer
  vision}, pages 764--773, 2017.

\bibitem{dai2021dynamic}
Xiyang Dai, Yinpeng Chen, Bin Xiao, Dongdong Chen, Mengchen Liu, Lu~Yuan, and
  Lei Zhang.
\newblock Dynamic head: Unifying object detection heads with attentions.
\newblock In {\em Proceedings of the IEEE/CVF Conference on Computer Vision and
  Pattern Recognition}, pages 7373--7382, 2021.

\bibitem{deng2009imagenet}
Jia Deng, Wei Dong, Richard Socher, Li-Jia Li, Kai Li, and Li~Fei-Fei.
\newblock Imagenet: A large-scale hierarchical image database.
\newblock In {\em 2009 IEEE conference on computer vision and pattern
  recognition}, pages 248--255. Ieee, 2009.

\bibitem{feng2021tood}
Chengjian Feng, Yujie Zhong, Yu~Gao, Matthew~R Scott, and Weilin Huang.
\newblock Tood: Task-aligned one-stage object detection.
\newblock In {\em Proceedings of the IEEE/CVF International Conference on
  Computer Vision}, pages 3510--3519, 2021.

\bibitem{gao2021fast}
Peng Gao, Minghang Zheng, Xiaogang Wang, Jifeng Dai, and Hongsheng Li.
\newblock Fast convergence of detr with spatially modulated co-attention.
\newblock In {\em Proceedings of the IEEE/CVF International Conference on
  Computer Vision}, pages 3621--3630, 2021.

\bibitem{ge2021ota}
Zheng Ge, Songtao Liu, Zeming Li, Osamu Yoshie, and Jian Sun.
\newblock Ota: Optimal transport assignment for object detection.
\newblock In {\em Proceedings of the IEEE/CVF Conference on Computer Vision and
  Pattern Recognition}, pages 303--312, 2021.

\bibitem{he2016deep}
Kaiming He, Xiangyu Zhang, Shaoqing Ren, and Jian Sun.
\newblock Deep residual learning for image recognition.
\newblock In {\em Proceedings of the IEEE conference on computer vision and
  pattern recognition}, pages 770--778, 2016.

\bibitem{kong2020foveabox}
Tao Kong, Fuchun Sun, Huaping Liu, Yuning Jiang, Lei Li, and Jianbo Shi.
\newblock Foveabox: Beyound anchor-based object detection.
\newblock {\em IEEE Transactions on Image Processing}, 29:7389--7398, 2020.

\bibitem{law2018cornernet}
Hei Law and Jia Deng.
\newblock Cornernet: Detecting objects as paired keypoints.
\newblock In {\em Proceedings of the European conference on computer vision
  (ECCV)}, pages 734--750, 2018.

\bibitem{li2022dn}
Feng Li, Hao Zhang, Shilong Liu, Jian Guo, Lionel~M Ni, and Lei Zhang.
\newblock Dn-detr: Accelerate detr training by introducing query denoising.
\newblock In {\em Proceedings of the IEEE/CVF Conference on Computer Vision and
  Pattern Recognition}, pages 13619--13627, 2022.

\bibitem{lin2017feature}
Tsung-Yi Lin, Piotr Doll{\'a}r, Ross Girshick, Kaiming He, Bharath Hariharan,
  and Serge Belongie.
\newblock Feature pyramid networks for object detection.
\newblock In {\em Proceedings of the IEEE conference on computer vision and
  pattern recognition}, pages 2117--2125, 2017.

\bibitem{lin2017focal}
Tsung-Yi Lin, Priya Goyal, Ross Girshick, Kaiming He, and Piotr Doll{\'a}r.
\newblock Focal loss for dense object detection.
\newblock In {\em Proceedings of the IEEE international conference on computer
  vision}, pages 2980--2988, 2017.

\bibitem{lin2014microsoft}
Tsung-Yi Lin, Michael Maire, Serge Belongie, James Hays, Pietro Perona, Deva
  Ramanan, Piotr Doll{\'a}r, and C~Lawrence Zitnick.
\newblock Microsoft coco: Common objects in context.
\newblock In {\em European conference on computer vision}, pages 740--755.
  Springer, 2014.
\newblock URL: \url{https://cocodataset.org}.

\bibitem{liu2022dab}
Shilong Liu, Feng Li, Hao Zhang, Xiao Yang, Xianbiao Qi, Hang Su, Jun Zhu, and
  Lei Zhang.
\newblock Dab-detr: Dynamic anchor boxes are better queries for detr.
\newblock {\em arXiv preprint arXiv:2201.12329}, 2022.

\bibitem{liu2018path}
Shu Liu, Lu~Qi, Haifang Qin, Jianping Shi, and Jiaya Jia.
\newblock Path aggregation network for instance segmentation.
\newblock In {\em Proceedings of the IEEE conference on computer vision and
  pattern recognition}, pages 8759--8768, 2018.

\bibitem{liu2016ssd}
Wei Liu, Dragomir Anguelov, Dumitru Erhan, Christian Szegedy, Scott Reed,
  Cheng-Yang Fu, and Alexander~C Berg.
\newblock Ssd: Single shot multibox detector.
\newblock In {\em European conference on computer vision}, pages 21--37.
  Springer, 2016.

\bibitem{loshchilov2017decoupled}
Ilya Loshchilov and Frank Hutter.
\newblock Decoupled weight decay regularization.
\newblock {\em arXiv preprint arXiv:1711.05101}, 2017.

\bibitem{meng2021conditional}
Depu Meng, Xiaokang Chen, Zejia Fan, Gang Zeng, Houqiang Li, Yuhui Yuan, Lei
  Sun, and Jingdong Wang.
\newblock Conditional detr for fast training convergence.
\newblock In {\em Proceedings of the IEEE/CVF International Conference on
  Computer Vision}, pages 3651--3660, 2021.

\bibitem{picron2021trident}
C{\'e}dric Picron and Tinne Tuytelaars.
\newblock Trident pyramid networks: The importance of processing at the feature
  pyramid level for better object detection.
\newblock {\em arXiv preprint arXiv:2110.04004}, 2021.

\bibitem{radosavovic2020designing}
Ilija Radosavovic, Raj~Prateek Kosaraju, Ross Girshick, Kaiming He, and Piotr
  Doll{\'a}r.
\newblock Designing network design spaces.
\newblock In {\em Proceedings of the IEEE/CVF Conference on Computer Vision and
  Pattern Recognition}, pages 10428--10436, 2020.

\bibitem{redmon2016you}
Joseph Redmon, Santosh Divvala, Ross Girshick, and Ali Farhadi.
\newblock You only look once: Unified, real-time object detection.
\newblock In {\em Proceedings of the IEEE conference on computer vision and
  pattern recognition}, pages 779--788, 2016.

\bibitem{ren2015faster}
Shaoqing Ren, Kaiming He, Ross Girshick, and Jian Sun.
\newblock Faster r-cnn: Towards real-time object detection with region proposal
  networks.
\newblock {\em Advances in neural information processing systems}, 28:91--99,
  2015.

\bibitem{rezatofighi2019generalized}
Hamid Rezatofighi, Nathan Tsoi, JunYoung Gwak, Amir Sadeghian, Ian Reid, and
  Silvio Savarese.
\newblock Generalized intersection over union: A metric and a loss for bounding
  box regression.
\newblock In {\em Proceedings of the IEEE/CVF conference on computer vision and
  pattern recognition}, pages 658--666, 2019.

\bibitem{sun2021sparse}
Peize Sun, Rufeng Zhang, Yi~Jiang, Tao Kong, Chenfeng Xu, Wei Zhan, Masayoshi
  Tomizuka, Lei Li, Zehuan Yuan, Changhu Wang, et~al.
\newblock Sparse r-cnn: End-to-end object detection with learnable proposals.
\newblock In {\em Proceedings of the IEEE/CVF Conference on Computer Vision and
  Pattern Recognition}, pages 14454--14463, 2021.

\bibitem{tan2020efficientdet}
Mingxing Tan, Ruoming Pang, and Quoc~V Le.
\newblock Efficientdet: Scalable and efficient object detection.
\newblock In {\em Proceedings of the IEEE/CVF conference on computer vision and
  pattern recognition}, pages 10781--10790, 2020.

\bibitem{tian2019fcos}
Zhi Tian, Chunhua Shen, Hao Chen, and Tong He.
\newblock Fcos: Fully convolutional one-stage object detection.
\newblock In {\em Proceedings of the IEEE/CVF international conference on
  computer vision}, pages 9627--9636, 2019.

\bibitem{vaswani2017attention}
Ashish Vaswani, Noam Shazeer, Niki Parmar, Jakob Uszkoreit, Llion Jones,
  Aidan~N Gomez, {\L}ukasz Kaiser, and Illia Polosukhin.
\newblock Attention is all you need.
\newblock In {\em Advances in neural information processing systems}, pages
  5998--6008, 2017.

\bibitem{wang2022anchor}
Yingming Wang, Xiangyu Zhang, Tong Yang, and Jian Sun.
\newblock Anchor detr: Query design for transformer-based detector.
\newblock In {\em Proceedings of the AAAI Conference on Artificial
  Intelligence}, volume~36, pages 2567--2575, 2022.

\bibitem{xie2017aggregated}
Saining Xie, Ross Girshick, Piotr Doll{\'a}r, Zhuowen Tu, and Kaiming He.
\newblock Aggregated residual transformations for deep neural networks.
\newblock In {\em Proceedings of the IEEE conference on computer vision and
  pattern recognition}, pages 1492--1500, 2017.

\bibitem{zhang2022dino}
Hao Zhang, Feng Li, Shilong Liu, Lei Zhang, Hang Su, Jun Zhu, Lionel~M Ni, and
  Heung-Yeung Shum.
\newblock Dino: Detr with improved denoising anchor boxes for end-to-end object
  detection.
\newblock {\em arXiv preprint arXiv:2203.03605}, 2022.

\bibitem{zhang2020dynamic}
Hongkai Zhang, Hong Chang, Bingpeng Ma, Naiyan Wang, and Xilin Chen.
\newblock Dynamic r-cnn: Towards high quality object detection via dynamic
  training.
\newblock In {\em European Conference on Computer Vision}, pages 260--275.
  Springer, 2020.

\bibitem{zhang2020bridging}
Shifeng Zhang, Cheng Chi, Yongqiang Yao, Zhen Lei, and Stan~Z Li.
\newblock Bridging the gap between anchor-based and anchor-free detection via
  adaptive training sample selection.
\newblock In {\em Proceedings of the IEEE/CVF conference on computer vision and
  pattern recognition}, pages 9759--9768, 2020.

\bibitem{zhou2019objects}
Xingyi Zhou, Dequan Wang, and Philipp Kr{\"a}henb{\"u}hl.
\newblock Objects as points.
\newblock {\em arXiv preprint arXiv:1904.07850}, 2019.

\bibitem{zhu2019deformable}
Xizhou Zhu, Han Hu, Stephen Lin, and Jifeng Dai.
\newblock Deformable convnets v2: More deformable, better results.
\newblock In {\em Proceedings of the IEEE/CVF Conference on Computer Vision and
  Pattern Recognition}, pages 9308--9316, 2019.

\bibitem{zhu2020deformable}
Xizhou Zhu, Weijie Su, Lewei Lu, Bin Li, Xiaogang Wang, and Jifeng Dai.
\newblock Deformable detr: Deformable transformers for end-to-end object
  detection.
\newblock {\em arXiv preprint arXiv:2010.04159}, 2020.

\end{thebibliography}
\bibliographystyle{plainurl}

\newpage
\appendix
\section{Revisiting DETR and Deformable DETR} \label{sec:revisit}

\paragraph{DETR.} DETR \cite{carion2020end} solves the object detection task using transformers \cite{vaswani2017attention}, originally designed to solve natural language processing (NLP) tasks. By doing so, DETR removes many hand-crafted components such as anchor generation and non-maximum suppression (NMS). It however comes at a cost. DETR needs more than 100 epochs to converge, which are many more than the dozen of epochs traditional detectors such as Faster R-CNN \cite{ren2015faster} need. Additionally, DETR does not perform well on small objects, as the quadratic complexity of self-attention layers prohibit the use of feature pyramids \cite{lin2017feature} inside the transformer encoder.

\paragraph{Vanilla Deformable DETR.} Not much later, Deformable DETR \cite{zhu2020deformable} is proposed. Deformable DETR is a DETR-like object detector mitigating the two main shortcomings of DETR, namely the slow convergence and the poor performance on small objects. Most of Deformable DETR's improvements can be attributed to the replacement of vanilla self-attention and cross-attention operations with new self-attention and cross-attention operations based on \textit{multi-scale deformable attention} (MSDA). MSDA is an operation extending deformable convolutions \cite{dai2017deformable, zhu2019deformable} using weights obtained via the softmax function as done in the attention-based literature \cite{vaswani2017attention}. Fundamentally, MSDA remains a local operation like traditional convolutions, in contrast to the global attention-based operations. By leveraging the local priors from MSDA, Deformable DETR was able to significantly speed up the convergence compared to DETR. Additionally, MSDA is \textit{multi-scale}, meaning that it can operate on feature pyramids (\ie a set feature maps of different scales). This allowed Deformable DETR to use feature pyramids at the encoder level and for cross-attention at the decoder level, considerably improving the performance on small objects.

\paragraph{Two-stage Deformable DETR.} In DETR-like detectors, the information of objects is captured in object queries or queries for short. Queries are updated within the decoder using cross-attention with the encoder output, self-attention between other queries, and feedforward computation on themselves. In DETR and vanilla Deformable DETR, these queries are initialized independently of the content of the image. This is sub-optimal as these queries should focus on regions containing objects, which could be inferred from the encoder output. Two-stage Deformable DETR addresses this issue by selecting features from its encoder output (\ie a feature pyramid) as initialization for the queries, further boosting its performance and convergence speed compared to vanilla Deformable DETR.

\end{document}